\documentclass[10pt,twocolumn,letterpaper]{article}

\usepackage[]{cvpr}  
\usepackage[ruled,linesnumbered]{algorithm2e}
\usepackage{algorithmic}
\usepackage{multirow}

\usepackage{booktabs}
\usepackage{wrapfig}
\usepackage{enumitem}

\usepackage[dvipsnames]{xcolor}

\definecolor{cvprblue}{rgb}{0.21,0.49,0.74}
\usepackage[pagebackref,breaklinks,colorlinks,citecolor=cvprblue]{hyperref}
\usepackage{graphicx}
\def\paperID{**} 
\def\confName{CVPR}
\def\confYear{2024}

\title{ART3D: 3D Gaussian Splatting for Text-Guided Artistic Scenes Generation}

\author{Pengzhi Li$^1$, Chengshuai Tang$^1$, Qinxuan Huang$^2$, Zhiheng Li$^{1{\dag}}$\\
$^1$Tsinghua Shenzhen International Graduate School \\ \
$^2$Tsinghua-Berkeley Shenzhen Institute \ \ 
}

\begin{document}

\twocolumn[{
\renewcommand\twocolumn[1][]{#1}
\maketitle
\begin{center}
    \captionsetup{type=figure}
    	\includegraphics[width=0.99\linewidth]{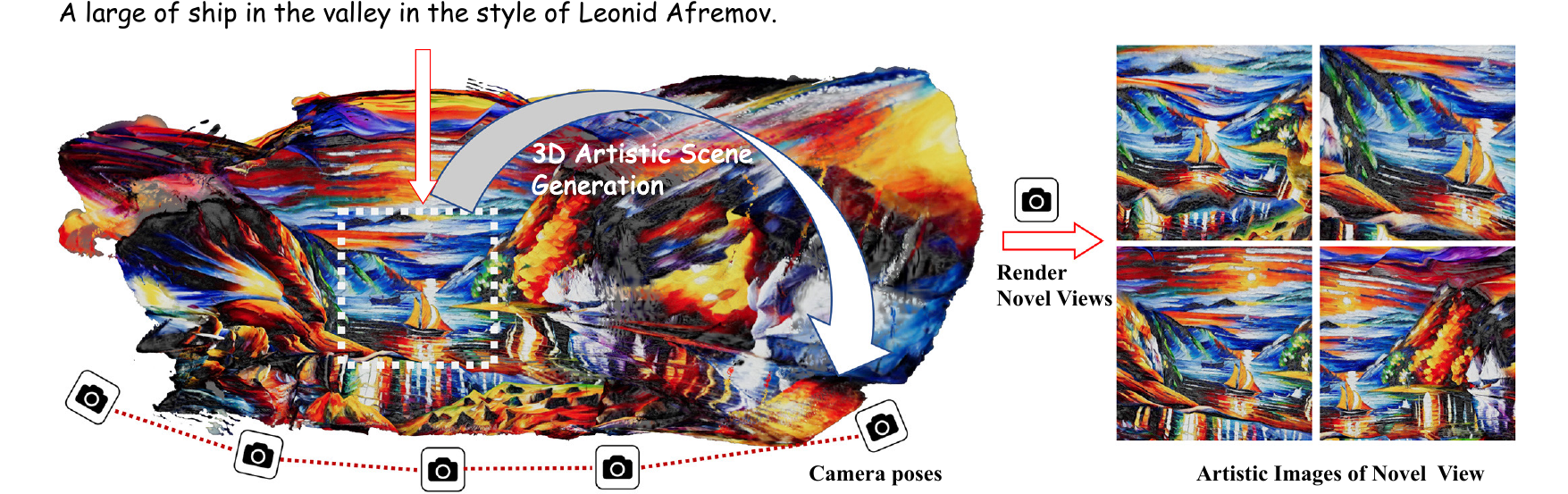}
\vspace{1em}
	\caption{We introduce ART3D, a novel framework for generating 3D artistic scenes. We can render 3D images with consistent views from predefined camera trajectories by inputting text descriptions. ART3D harnesses the powerful generation capabilities of the diffusion model and state-of-the-art 3D Gaussian splatting, fostering advancements in AI-driven art creation.}
	\label{fig:teaser}
 \vspace{1em}
\end{center}
}]

\let\thefootnote\relax\footnotetext{
$^\dag$Corresponding author.
}
\maketitle
\vspace{1em}
\begin{abstract}

In this paper, we explore the existing challenges in 3D artistic scene generation by introducing ART3D, a novel framework that combines diffusion models and 3D Gaussian splatting techniques. Our method effectively bridges the gap between artistic and realistic images through an innovative image semantic transfer algorithm. By leveraging depth information and an initial artistic image, we generate a point cloud map, addressing domain differences. Additionally, we propose a depth consistency module to enhance 3D scene consistency. Finally, the 3D scene serves as initial points for optimizing Gaussian splats. 
Experimental results demonstrate ART3D's superior performance in both content and structural consistency metrics when compared to existing methods. ART3D significantly advances the field of AI in art creation by providing an innovative solution for generating high-quality 3D artistic scenes.

\end{abstract}

\section{Introduction}

With the progress of Artificial Intelligence Generated Content (AIGC) and advancements in 3D vision technology, AI-driven art creation~\cite{jiang2023aiart_1,guo2023can_art2,miah2023generativeart} has become a hot topic for designers and artists. The latest generative models~\cite{rombach2022high,ho2022imagen} in the realm of 2D art are capable of producing high-quality images based on artistic prompts, making it easier for users to create 2D art and promoting the growth of AI art. However, despite success in the 2D field, 3D art creation still faces significant challenges.

The current approaches face a series of challenges when attempting to apply similar generative models to create 3D models from textual descriptions. While some emerging methods~\cite{poole2022dreamfusion,raj2023dreambooth3d,zhao2023efficientdreamer} strive to address the issues from text to 3D models, significant difficulties arise due to the lack of available 3D art training data~\cite{chen2019text2shape,bautista2022gaudi}, especially data tailored to artistic creations rather than real-world scenes. These methods encounter substantial challenges when dealing with domain differences between real-world scenes and artistic works. Furthermore, some approaches~\cite{shih20203d,liu2021infinite,hollein2023text2room} adopt a two-step process of generating images first and then converting them to 3D. However, the 3D information required by these models is often based on training with real-world datasets, making accurate information prediction for artistic inputs a difficult task. All these challenges contribute to the fact that research on 3D technology in artistic creation is still in its early exploratory stages.

In this paper, we aim to explore a novel research question: how to generate high-quality 3D scenes with artistic styles based on text or reference RGB images. As shown in Figure 1, leveraging the powerful generative capabilities of Stable Diffusion models~\cite{rombach2022high}, we introduce an innovative 3D Gaussian splatting~\cite{kerbl20233dGS} technique to produce artistic 3D scenes from user-provided textual input by updating a point cloud map.

Specifically, we propose an image semantic transfer algorithm to bridge the gap between artistic and realistic images. This algorithm extracts feature maps with the same semantic layout obtained through UNet from user-provided text or reference images. Subsequently, we use text prompts to generate new realistic images, ensuring they adhere to the same semantic layout. While realistic images play a role in the intermediate layers, once we predict the depth information of realistic images, we can transfer this information to their corresponding artistic image. Using the depth map and the initial artistic image to generate an initial point cloud, we successfully address the domain gap between artistic and realistic images.

Next, we generate a reasonable intrinsic matrix and camera pose, using the camera pose to reproject the initial point cloud onto a new camera plane. Employing the inpainting technique of the Stable Diffusion model~\cite{rombach2022high}, we complete the hollow areas of the novel view image, ultimately obtaining a complete image. By iteratively performing point cloud generation and projection, we finally obtain an initial point for the scene point cloud optimized using 3D Gaussian splatting~\cite{kerbl20233dGS}. In this process, we introduce a depth consistency module to enhance consistency between multiple views. Aligning depth domains, this module ensures seamless integration of the newly generated point cloud into the entire system, improving consistency in multi-view generation. Ultimately, we render an impressive 3D scene based on a continuous representation of the point cloud map using 3D Gaussian splatting.

Our experiments demonstrate that our approach excels in both style consistency and continuity metrics for artistic images compared to existing methods, exhibiting greater visual appeal. We further include the ablation studies involving different core components. Finally, we extend our method to various application scenarios, such as text-driven fusion of different artistic styles. This paper provides an innovative approach to high-quality artistic 3D scene generation based on textual descriptions or reference images, making significant contributions to the intersection of art and technology. Our contributions can be summarized as follows:

\begin{itemize}
	\item We introduce ART3D, which achieves high-quality 3D artistic scene generation through diffusion models and 3D Gaussian splatting techniques.
 
	\item Our method compensates for the domain gap between artistic and realistic images through an image semantic transfer algorithm, and the introduction of a depth consistency module improves the overall consistency of global scene generation.
 
	\item We innovatively address the generation of high-quality 3D artistic scenes from text or reference images, making a significant contribution to the development of the interdisciplinary field of AI in art creation.
\end{itemize}

\begin{figure*}[t]
	\centering

	\includegraphics[width=0.82\linewidth]{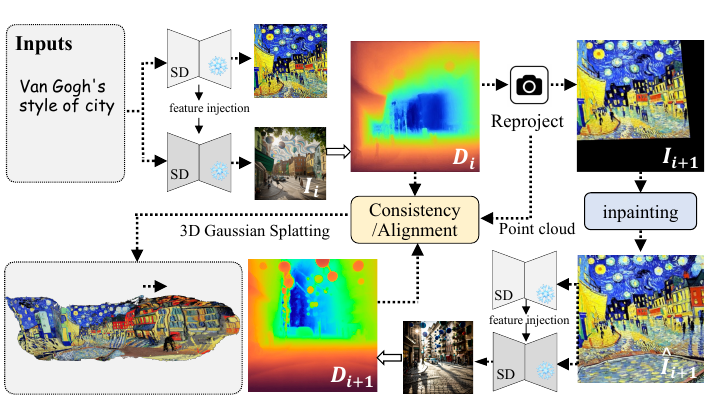}

	\caption{\textbf{Pipeline of ART3D. }We introduce the process of our method. First, we input textual descriptions or reference images and use semantic transfer algorithms to obtain accurate depth 3D information. Then, construct a point cloud and enhance multi-view consistency through consistency/alignment algorithms. Finally, we render high-quality 3D artistic scenes using 3D Gaussian splatting technology.}
	\label{fig:pipeline_0}
\end{figure*}
\section{Related Works}
\label{sec:relate}

In this section, we provide a brief review of the literature related to diffusion models, text-guided 3D generation and 3D scene synthesis.

\noindent\textbf{Diffusion Models.}
In recent years, large-scale models have become a hot topic in the field of AI. Among them, diffusion models~\cite{ho2020denoising,rombach2022high} are widely recognized as powerful tools in the areas of complex data modeling and generation. Their remarkable and stable capabilities in complex data modeling have led to successful applications in various domains, including image restoration~\cite{lugmayr2022repaint}, image translation~\cite{saharia2022palette}, image /video generation~\cite{khachatryan2023text2video,li2023archi}, super-resolution~\cite{saharia2022image}, and image editing~\cite{meng2021sdedit,li2023layerdiffusion,li2024tuning}. P2P~\cite{hertz2022prompt} and PnP~\cite{tumanyan2023plug} further explore the crucial role of attention mechanisms and features in the process of image generation, proposing the use of cross-channel attention or self-attention mechanisms for global image editing tasks. Recent research indicates that these techniques have successfully extended from the 2D domain to the 3D domain, as demonstrated in~\cite{luo2021diffusion,poole2022dreamfusion}.
Our approach maximizes the advantages of diffusion models in both 2D and 3D domains, leveraging their powerful restoration and generation capabilities, as well as feature mechanisms. This has led to successful applications in the field of AI artistic creation.

\noindent\textbf{Text-to-3D.}
With the emergence of pre-trained text-conditioned image generation diffusion models, the trend of extending these 2D models to achieve text-to-3D generation has become increasingly popular. Some methods employ supervised training of text-to-3D models using 3D data~\cite{chen2019text2shape,bautista2022gaudi,zhou20213d}; however, due to the lack of large-scale aligned text and 3D datasets, this direction remains challenging. To achieve open-vocabulary 3D generation, pioneering work by DreamFusion~\cite{poole2022dreamfusion} has paved the way, where some approaches propose the application of text-to-image diffusion models~\cite{rombach2022high}to 3D generation~\cite{wang2023prolificdreamer,zhao2023efficientdreamer,li2024generating,chan2023generative,tang2023make}. Subsequently, numerous extensive improvements have been made to DreamFusion~\cite{lin2023magic3d,raj2023dreambooth3d,tang2023dreamgaussian,wang2023prolificdreamer}. ProlificDreamer~\cite{wang2023prolificdreamer} introduces Variational Score Distillation (VSD) and generates high-fidelity texture results. Magic3D~\cite{lin2023magic3d} adopts a coarse-to-fine strategy and utilizes DMTET~\cite{shen2021deep} as a 3D representation to achieve texture refinement through SDS loss. Recent 
methods~\cite{lin2023magic3d,melas2023realfusion,metzer2023latent,wang2023score} combine large-scale text-to-image diffusion models~\cite{saharia2022photorealistic,rombach2022high} with neural radiance fields~\cite{mildenhall2021nerf} to generate 3D objects without the need for training. However, most of these methods are tailored to real images or common stylistic works. Approaches for text-driven 3D scene generation, particularly for diverse artistic styles, remain largely unexplored.

\noindent\textbf{3D Scene Synthesis.} 
The typical methods for presenting 3D scenes include explicit approaches like point clouds, meshes, and voxels. Recently, there has been a focus on implicit methods, such as signed distance functions~\cite{park2019deepsdf,park2019deepsdf} and neural radiance fields~\cite{mildenhall2021nerf}, for expressing 3D scenes. The 3D Gaussian splatting technique~\cite{kerbl20233dGS}, by cleverly combining Gaussian splats, spherical harmonics, and opacity, achieves a fast and high-quality reconstruction of complete and boundaryless 3D scenes. Its features align well with our approach to freely generate scenes, so we opt to use this rendering method in this paper.

\newcommand{\bs}[1]{{\boldsymbol{#1}}}

\section{Method}

In this section, we provide a detailed overview of ART3D, which consists of four key components, as illustrated in Figure 2. Firstly, we leverage the attention mechanism of Stable Diffusion models~\cite{rombach2022high} and design an image semantic transfer algorithm to enhance the accuracy of obtaining depth information from artistic images. In Section 3.2, we establish a point cloud map that transforms depth information $D_{i}$ into a point cloud, and through camera reprojection, generates new images $I_{i+1}$. Subsequently, we apply the operations from the first part to the inpainting image $\hat{I}_{i+1}$ to obtain depth map $D_{i+1}$. In Section 3.3, we introduce a depth consistency module to improve consistency between multiple views and seamlessly integrate the new point cloud into the existing point cloud map. Finally, employing the 3D Gaussian splatting technique, we successfully render high-quality 3D artistic scenes and novel views.

\subsection{Image Semantic Transfer}
\label{sec:twin}

The pre-trained stable diffusion model~\cite{rombach2022high} can generate an artistic image with a corresponding style based on input prompts such as mosaic, Van Gogh, etc. However, simply appending keywords of realistic styles at the end of the prompts makes it nearly impractical to generate images that align with semantic information due to the stochastic nature of the diffusion model. Currently, conditional generation methods like ControlNet~\cite{zhang2023adding}, which leverage depth and edge information as conditional inputs, struggle when applied to artistic images, as these artistic images inherently possess fuzzy characteristics. Therefore, adopting such an approach is still challenging.

To address this issue and generate real images with a semantic layout similar to artistic images, similar to~\cite{mahapatra2023text}, we utilize the internal features of the Stable Diffusion model to align the semantic information of the two images. The intermediate layers of the UNet in the Stable Diffusion model consist of residual blocks, self-attention layers, and cross-attention layers. The self-attention layer of the $i$-th block at time $t$ computes features as follows:

\begin{equation}
    \bs{f}_{t}\xspace = \bs{A}^i_{t}\bs{V}_{t}^i,  \bs{A}_{t}^i = \text{Softmax}(\dfrac{\bs{Q}^i_{t}\bs{K}_{t}^{i^T}}{\sqrt{d}}),
\end{equation}
where $\bs{Q}^i_{t}, \bs{K}^i_{t}, \bs{V}^i_{t}$ 
represent the query, key, and value.

Similar to PnP~\cite{tumanyan2023plug}, we note that the self-attention maps $\bs{A}_t$ at each time step $t$ during the denoising process in Stable Diffusion control the spatial structure of the resulting image. Therefore, while generating the artistic image $\bs{x}\xspace$, we retain the intermediate self-attention maps $\bs{A}_t$ for all time steps $t$. Injecting the output residual block features $\bs{f}_{t}\xspace$ is employed to enhance structural alignment. Specifically, we store the features from the 4-th layer in the output blocks. Then, we use the self-attention maps during the denoising process of Stable Diffusion to control the layout of the generated image. We inject the output residual block features $\bs{f}_{t}\xspace$ and self-attention maps $\bs{A}_t\xspace$ into the UNet module. This process can be represented as follows:

\begin{align}
&\begin{aligned}
\bs{x}_{t-1}\xspace, \bs{A}_t\xspace, \bs{f}_{t}\xspace &= \bs{\epsilon_\theta}(\bs{x}_t\xspace, \bs{c}\xspace, t),
\end{aligned} \ 
\\
&\begin{aligned}
\hat{\bs{x}}_{t-1}\xspace &= \bs{\hat{\epsilon}_\theta}(\hat{\bs{x}}_t\xspace, \hat{\bs{c}}\xspace, t; \bs{A}_t\xspace, \bs{f}_{t}\xspace),
\end{aligned}
\end{align}

where $\bs{\epsilon_\theta}$ is a standard denoising UNet. $\bs{x}_t\xspace$ and $\hat{\bs{x}}_t\xspace$ are noisy images at time step $t$ corresponding to the artistic and realistic images, respectively. $\bs{\hat{\epsilon}_\theta}$ is the modified UNet, which takes injected features as input.

During the generation process, the alignment of semantic features between artistic and realistic images is guaranteed by leveraging the stored features $\bs{f}_{t}\xspace$ and self-attention maps $\bs{A}_t\xspace$.

\subsection{Point Cloud Map}
\label{sec:pc}

Our approach achieves view-consistent 3D scene generation by updating a point cloud map. Initially, we set up the camera's intrinsic and extrinsic parameters as $T$. Next, utilizing a monocular scale depth estimation model, ZoeDepth~\cite{bhat2023zoedepth}, we acquire a depth map $D_{i}$ and lift pixels of $I_{i}$ to 3D space based on depth information, thus creating an initial point cloud $P_{i}$. Along the camera's trajectory, we project the point cloud $P_{i}$ and then obtain a new image $I_{i+1}$ by reprojecting it back onto the camera plane. This process can be represented as follows:

\begin{equation}
    P_{i} = {\mathbf{\phi}}(I_{i},D_{i},T),
    I_{i+1} = {\mathbf{\psi}}(P_{i},T),
    \label{eq:01}
\end{equation}
where ${\mathbf{\phi}}$ and ${\mathbf{\psi}}$ denote the projection from 2D to 3D and the reprojection from 3D to 2D.

Subsequently, we utilize a Stable Diffusion model~\cite{rombach2022high} to inpainting the missing areas in $I_{i+1}$. We then repeat the aforementioned process ${\mathbf{\phi}}$ and ${\mathbf{\psi}}$, obtaining scene images from every viewpoint along the camera's trajectory. Due to the independence among these 3D point clouds and the discontinuity in the depth estimation model, the scale information among these point clouds varies. Therefore, in Section 3.3, we introduce an efficient depth consistency algorithm to standardize the initial depth scale information obtained from the depth estimation model, ensuring the maintenance of a scale-consistent point cloud map. 

Next, it is essential to align the projected point clouds corresponding to each pose to ensure the generation of a continuous 3D artistic style scene. The 3D point cloud $P_{i}$, projected from the camera pose of the previous time step, results in the image $I_{i+1}$. After inpainting, a new image $\hat{I}_{i+1}$ (representing the point cloud to be aligned) is obtained from $I_{i+1}$. In $\hat{I}_{i+1}$, a significant portion of content overlaps with $I_{i}$, and we denote this overlapping region as the $M$. Consequently, the 3D point clouds $P_{i+1}$ and $P_{i}$, projected from the current pose of $\hat{I}_{i+1}$, also share many redundant points. The point alignment can be represented as follows:
\begin{equation}
\min_{s} \|M \times ({\mathbf{\phi}}(I_{i} ,S \times D_{i},T) - P_{i+1})\|_1.
    \label{eq:02}
\end{equation}
We obtain the aligned coarse global factor $S$ by optimizing the two point cloud regions using $M$. Due to our earlier depth consistency module, the aligned point cloud maintains global and local consistency.

\begin{figure}[t]
	\centering

	\includegraphics[width=\linewidth]{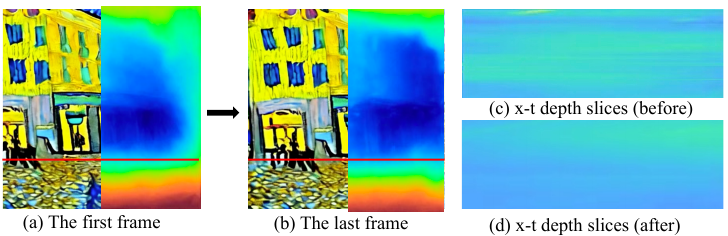}

	\caption{We demonstrate the effectiveness of our depth consistency module. Frames from (a) to (b) represent consecutive frames, while (c) and (d) illustrate depth value slices evolving over time at the red line in (a) and (b). A smoother x-t slice indicates more consistent depth. Our approach significantly enhances depth consistency.}
	\label{fig:depth consistency}
	
\end{figure}

\subsection{Depth Consistency Module}\label{sec:method_dcm}

According to the analysis in Section 3.2, achieving a consistent 3D point cloud scene necessitates relying on consistent depth information. Due to the independence of predictions from depth estimation networks, the depth consistency algorithm faces two challenges: the global distribution of depth ranges for different depths is inconsistent, making it impractical to simply employ a global scaling factor $S$ for standardization. Additionally, methods~\cite{luo2020consistent,li2023efficient,li2023towards} that involve the re-optimization of depth information based on camera poses require RGB domain image information, making their application challenging in the artistic domain.

To address these challenges, we propose DCM, which eliminates the dependence on RGB domain image information, thereby enhancing depth consistency across different views of the same 3D scene. Specifically, DCM takes the depth $D_{i}$ from the previous view and the initial depth $D_{i+1}$ from the current view as input, producing the updated depth $D_{i+1}$ as output. This can be represented as follows:
\begin{equation}
{D}_{i+1} ={D}_{i} + DCM\left ( {D}_{i+1}, {D}_{i} \right ).
\label{eq:dcm}
\end{equation}
Based on this formula, DCM attempts to learn a depth residual to update $D$ and minimize the inconsistency between $D_{i}$ and $D_{i+1}$. To ensure computational efficiency, we employ shallow convolutional layers and residual blocks to implement DCM. Its architecture comprises an encoder and a decoder, connected via skip connections.

While the structure of DCM is concise, we encounter a scarcity of real multi-view datasets containing ground truth depth from different perspectives, posing challenges to DCM's training. Considering that our scenario only involves static scenes without dynamic objects, we train the model using the virtual multi-frame dataset IRS~\cite{wang2021irs}. This dataset includes highly accurate depth information for static scenes. Specifically, we select a set of 7 frames as training data and the initial depth map is resized and cropped to a resolution of 384. To improve the depth accuracy, we also employ a depth domain loss similar to MiDaS~\cite{ranftl2020midas}, which constrains the same depth range. Additionally, we design a depth consistency loss as follows: 

\begin{equation}
\label{eq:tc_loss_update}
D_{i}\Leftarrow D_{i} + DCM(D_{i+1}, D_{i}),
\end{equation}
\begin{equation}
\label{eq:tc_loss}
\mathcal L_{C}=\sum_{i=1}^{7} \left\|M_{s}\left(D_{i+1}-\hat{D}_{i}\right)\right\|_1,
\end{equation}

where $M_{i}=\exp (-\alpha|O_{i+1}-\hat{O}_{i}|^{2})$ represents the occlusion weight between $D_{i+1}$ and $\hat{D}_{i}$, and $\alpha$ is set to $50$. 
$O_{i}$ and $O_{i+1}$ are RGB frames.
$\hat{D}_{i}$ is derived by warping $D_{i}$ to $D_{i+1}$ based on the backward optical flow $F_{i+1 \Rightarrow i}$, which is computed using GMFlow~\cite{xu2022gmflow}. Furthermore, $\hat{O}{i}$ is obtained by warping $O_{i}$ to $O_{i+1}$ according to $F_{i+1 \Rightarrow i}$. This involves utilizing the optical flow to transform visualized images, ensuring alignment between $O_{i}$ and $O_{i+1}$. It is important to note that $D_{i}$ is warped by the optical flow $F_{i+1 \Rightarrow i}$.

As shown in Figure 3, we demonstrate how our approach effectively enhances depth consistency, where (c) and (d) represent x-t slices of scene motion, with more smooth indicating better depth consistency.

\begin{figure*}[t]
	\centering

	\includegraphics[width=0.98\linewidth]{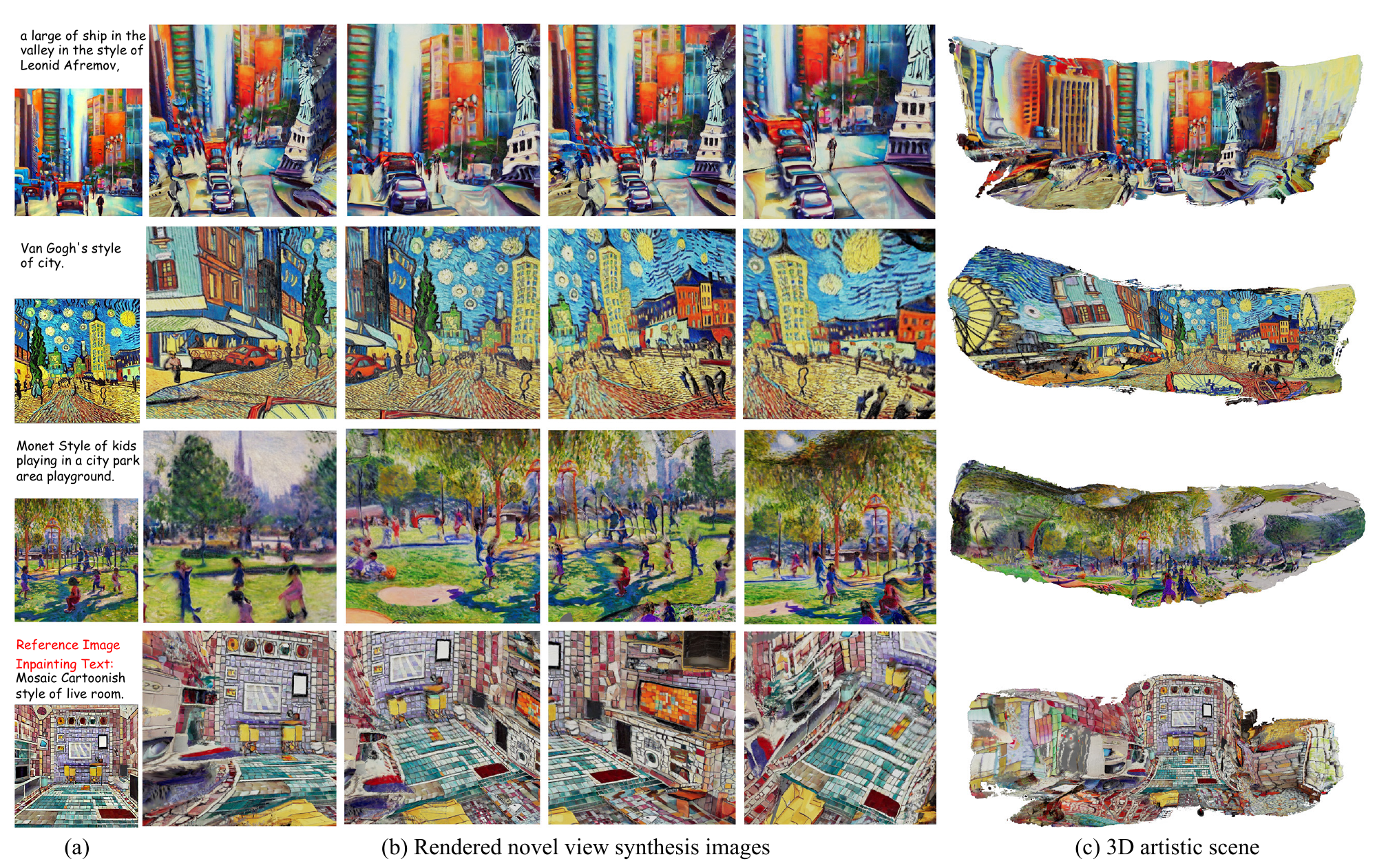}

	\caption{\textbf{Visualization results of our method.} (a) represents the inputs, where we can use only text or a combination of a reference image and text as input. (b) is a novel view image rendered from the generated 3D artistic scene, showing stylistic consistency. (c) demonstrates 3D artistic scenes generated by our method through a predefined camera trajectory. Our approach can accurately and high-quality generate structurally consistent and diverse 3D artistic scenes.}
	\label{fig:visualization}
\end{figure*}

\subsection{3D Gaussian Splatting for Rendering}
\label{sec:gaussdepth}

Following the point cloud alignment stage, we obtain a complete point cloud map, which serves as the initial point cloud for the Structure from Motion (SfM) required in training the 3D Gaussian splatting~\cite{kerbl20233dGS} model. Each Gaussian splats point is initialized with these point cloud values and optimized for volume and position using the ground truth from the projection images as supervision. In contrast to traditional SfM methods~\cite{schonberger2016structure}, which may exhibit shortcomings in the domain of artistic images, our approach not only yields more accurate initial point clouds but also accelerates the convergence of the network, facilitating deeper learning of more nuanced features. Additionally, considering that regions generated by the Stable Diffusion Inpainting model~\cite{rombach2022high} may contain inaccurate information, we intentionally ignore these areas when computing the loss function. Given that points in the point cloud are represented by Gaussian distributions, these regions will naturally be filled during the training process.

\begin{figure*}[t]
	\centering

	\includegraphics[width=\linewidth]{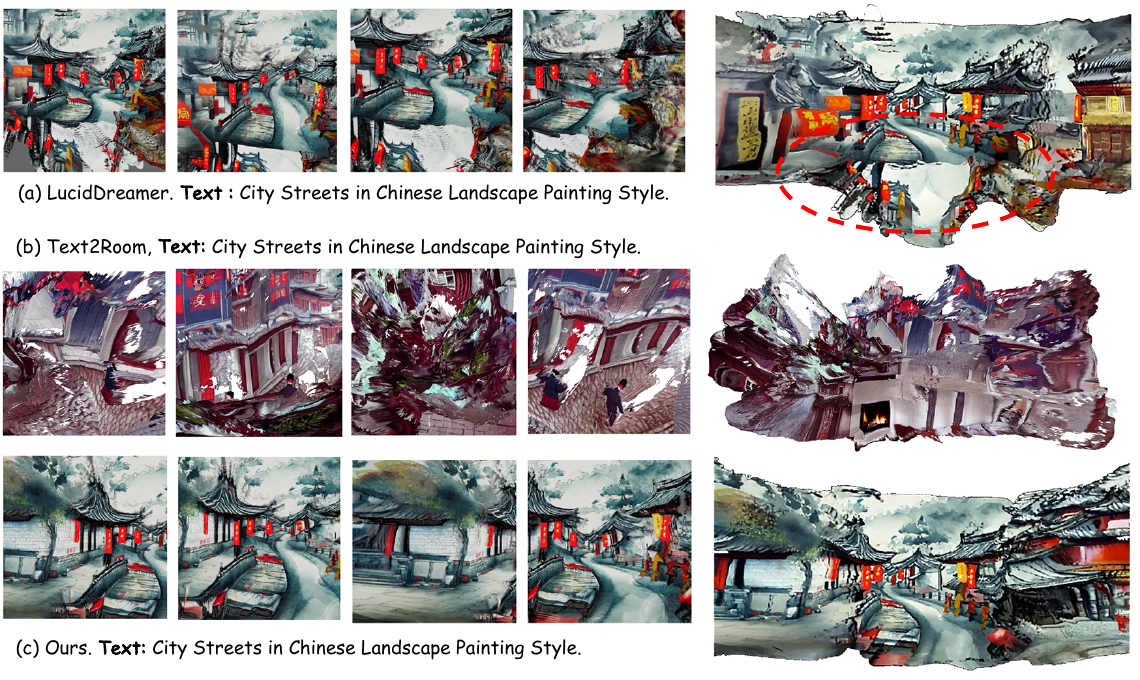}

	\caption{\textbf{Qualitative comparison with 3D scene generation methods. }LucidDreamer~\cite{chung2023luciddreamer} and Text2Room~\cite{hollein2023text2room} perform poorly on artistic images due to the gap between the artistic and realistic domains. As highlighted in the red circle, they struggle to obtain accurate 3D information, leading to structural errors caused by depth accuracy or alignment. These issues result in the generation of unsatisfactory 3D scenes. In contrast, our method excels in the artistic domain.}
	\label{fig:01}
\end{figure*}

\section{Experiments}

In this section, we present the details of our experiments, the results of comparisons, and qualitative results of ART3D.

\subsection{Experiment Setup}

\paragraph{Implementation Details}For our text-to-image model, we adopt the Stable Diffusion~\cite{rombach2022high} model and its inpainting version, fine-tuned on the image inpainting task with an additional mask input. Concurrently, we employ the ZoeDepth~\cite{bhat2023zoedepth} model as our monocular depth estimator, representing the SOTA in scale-depth estimation models. Subsequently, we provide a detailed overview of the training parameters for our DCM below.
\paragraph{Details of DCM} The DCM adopts an encoder-decoder architecture, where the encoder consists of two downsampling strided convolutional layers, followed by five residual blocks. In the decoding process, two transposed convolutional layers are employed, incorporating essential skip connections from the encoder to the decoder. Instance Normalization is consistently applied, with the exception of the last layer. To confine the output within the range of -1 to 1 after decoding, a Tanh layer is employed. The model is implemented using PyTorch and is trained with the Adam solver~\cite{kingma2014adam} for 20,000 iterations, maintaining a steady learning rate of 1e-4. Throughout the training phase, a batch size of 4 is utilized, and the training data undergoes random cropping to dimensions of 384×384.

\paragraph{Evaluation Metrics}
We need to ensure that the rendered images maintain consistency with the reference images' styles. To achieve this, we employ CLIP-I~\cite{radford2021learning} metrics to calculate the similarity of image features. The second metric needs to evaluate the consistency between the rendered image and textual descriptions. Additionally, we utilize the CLIP-T metric to calculate the cosine similarity between text prompts and CLIP embeddings. Furthermore, user studies are conducted to evaluate the feasibility of our approach comprehensively.

\begin{table}[h]
  \centering
  \resizebox{0.48\textwidth}{!}{
  \begin{tabular}{l cc cc}
    \toprule
        \multirow{2}{*}{Method} & \multicolumn{2}{c}{2D Metrics} & \multicolumn{2}{c}{User Study}\\
                        \cmidrule(l{2pt}r{2pt}){2-3} \cmidrule(l{2pt}r{2pt}){4-5}
    & CLIP-I $\rightarrow$ & CLIP-T $\rightarrow$ & SC $\rightarrow$ & CC $\rightarrow$\\
    \midrule
    Text2room~\cite{hollein2023text2room} & 53.44 & 21.32 & 2.78 & 2.53\\
    LucidDreamer~\cite{chung2023luciddreamer} & 64.43 & 25.87 & 3.65 & 4.11\\
    \midrule
    Ours w/o IST & 62.86 & 24.32 & 3.57 & 3.98\\
    Ours w/o PCM & 60.36 & 23.34 & 3.72 & 3.78\\
    Ours w/o DCM & 63.32 & 24.59 & 3.46 & 3.89\\
    Ours & \textbf{68.15} & \textbf{26.81} & \textbf{4.01} & \textbf{4.32}\\
    \bottomrule
  \end{tabular}
  }
  \caption{
  \textbf{Quantitative comparison.}
  We report image metrics and user study results, including ClIP-I and CLIP-T Score. We also use two metrics for user study: Structural Consistency Score and Content Consistency.
  Our method creates 3D artistic scenes with the highest quality.}
  \vspace{-1em}
  \label{tab:quantitive_result}
\end{table}

\begin{figure}[t]
	\centering

	\includegraphics[width=\linewidth]{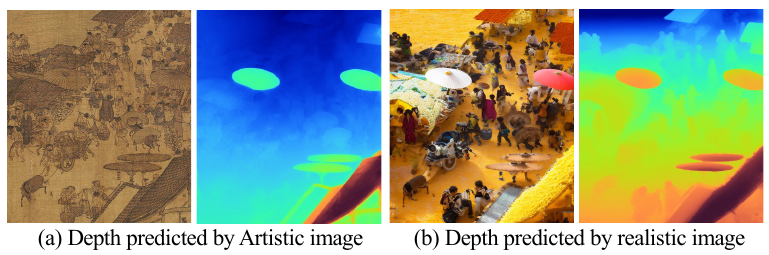}

	\caption{The depth map directly estimated from the artistic image (a) lacks numerous details. In (b), our image semantic transfer algorithm can generate a more accurate depth map. This addresses the challenge of obtaining precise 3D information to generate 3D artistic scenes.}
	\label{fig:depth_ablation}
	
\end{figure}

\begin{figure}[t]
	\centering

	\includegraphics[width=\linewidth]{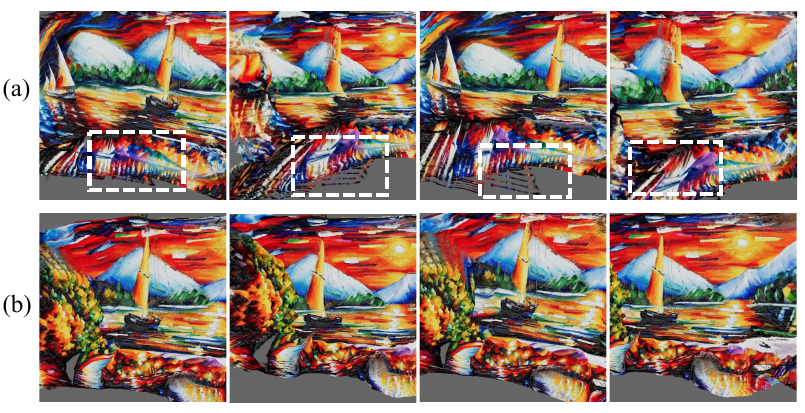}

	\caption{We demonstrated the effectiveness of our depth consistency module. In (a), we optimize only the global depth scale factor $S$, leading to discontinuities and structural distortions within the white region. After processing with our depth consistency algorithm, the generated 3D scenes exhibit improved consistency between different views.}
	\label{fig:pipeline}
	
\end{figure}

\subsection{ Qualitative Results}

As shown in Figure 4, we illustrate the results generated by our method when provide with multiple sets of artistic textual descriptions. Our ART3D successfully produces consistent 3D artistic scenes, as depicted in (d). Leveraging the generative capabilities of the diffusion model, our approach achieves stylistic coherence in scene generation based on textual prompts. Additionally, as shown in (b), we can render images from various perspectives by incorporating given camera poses, presenting a novel approach for novel view image synthesis tasks.

To further assess the superiority of our method in text-driven 3D artistic scene generation, we conduct a comparison with another method utilizing the diffusion model for 3D scene generation~\cite{chung2023luciddreamer,hollein2023text2room} (Figure 5). In contrast to approaches applied in real-world domains or those closely simulating real-world texture, our method produces more continuous multi-view images and more consistent and plausible 3D scenes. The structural consistency in our generated 3D scenes surpasses that of other methods, which may exhibit errors in spatial positioning due to depth alignment issues, as indicated within the red circles in Figure 5. Specifically, LucidDreamer~\cite{chung2023luciddreamer} exhibits some depth alignment problems, while Text2Room~\cite{hollein2023text2room} shows reduced robustness for outdoor scenes, leading to potential inaccuracies in structural information.

\subsection{Quantitative Results}

As shown in Table 1, we present the averaged quantitative results for multiple scenes. We generate ten distinct styles of 3D artistic scenes using textual descriptions and render 50 new scene images from each scenario for calculating image quality metrics. CLIP-I~\cite{radford2021learning} and CLIP-T scores are employed to assess the similarity between generated images, reference images, and textual descriptions. Simultaneously, we conduct a user study, selecting 25 sets of images for 20 users to rate (0 to 5 represents worst to best score). Specifically, we focus on structural consistency metrics(SC), such as the presence of incoherent elements in the scenes, such as holes, and content consistency metrics (CC), assessing the alignment of images with textual descriptions. Across all these metrics, our approach consistently achieves the highest scores, highlighting its ability to generate structurally consistent high-quality 3D artistic scenes accurately.

\section{Ablation Studies}

The key components of our method include the image semantic transfer algorithm (Section 3.1), point cloud map (Section 3.2), and depth consistency (Section 3.3). We conduct an extensive ablation study to validate the effectiveness of each component.

\paragraph{Effects of Image Semantic Transfer}

As we know, most depth estimation models are typically trained on real datasets, posing challenges in accurately estimating depth information for artistic images. However, accurate depth information is crucial for generating point clouds, ensuring the fidelity of the generated scenes. Our designed image semantic transfer algorithm can generate realistic scene images with the same semantic layout as corresponding artistic style images. As shown in Figure 6, directly predicted depth map (a) struggles to capture scene information accurately, while our method can generate a depth map (b) with rich details, leading to more accurate point clouds and avoiding issues like object misplacement. Our approach effectively bridges the gap between artistic and real-world image applications and provides new inspiration and insights for AI artistic creation.

\paragraph{Effects of Point Cloud Map}

We evaluate the point cloud generation strategy we employed for 3D Gaussian Splatting initialization. We compare it with point clouds obtained using COLMAP~\cite{schonberger2016structure}. For the task of artistic scene generation, our point cloud exhibits exceptionally high quality, providing a substantial number of high-quality points for 3D Gaussian Splatting initialization and significantly accelerating the reconstruction speed. As shown in Table 2, we evaluate our generated artistic results using image quality metrics, demonstrating the effectiveness of our method in enhancing the rendering quality of artistic scenes.

\paragraph{Effects of Depth Consistency Module}

Due to the fact that the 3D information required for point cloud generation is predicted from a monocular depth estimation model, inconsistencies in scale and depth range exist in the predictions of monocular depth across consecutive frames. Our proposed monocular depth consistency algorithm ensures the generation of a coherent 3D scene, thereby avoiding issues such as discontinuities, holes, gaps, and distortions in the final 3D representation. This algorithm learns residual information between depths from two different viewpoints, effectively unifying the depth range and scale. In Figure 7 (a), we present visual results where a simple global scaling factor for depth alignment leads to noticeable discontinuities and distortions in the 3D structure within the white boxes.

\begin{table}
  \small 
  \renewcommand{\arraystretch}{0.4}
  \centering
  \resizebox{0.45\textwidth}{!}{

    \begin{tabular}{lcc c c}
      \toprule
      Metric  &  Method  & 1000 iters& 4000 iters & 7000 iters\\
      \midrule
      PSNR $\uparrow$ & COLMAP   & 21.125& 22.287 & 23.138  \\
      & Ours  &\underline{23.032} & 23.613 & \textbf{24.041} \\
      \midrule
      SSIM $\uparrow$ &COLMAP  &0.810  & 0.834 & 0.843 \\
      & Ours & \underline{0.838} & 0.847 & \textbf{0.863} \\
      \midrule
      LPIPS  $\downarrow$&COLMAP  &0.258 & 0.237 & 0.224  \\
       & Ours&\underline{0.229} & 0.221 & \textbf{0.214}  \\
      \bottomrule
    \end{tabular}
  }
  \caption{
    \textbf{Quantitative comparison.}
    We compare our method with the 3D artistic scene reconstruction results obtained using COLMAP. \underline{Underline} shows that we achieve competitive performance with few iterations. Our approach optimizes 3D Gaussian splats more efficiently and exhibits superior reconstruction metrics.
  }\label{tab:colmap_result}
\end{table}

\section{Conclusion}

In conclusion, ART3D represents an advancement in AI-driven 3D art creation. By effectively addressing challenges in domain gaps and global scene consistency, our approach, utilizing diffusion models and 3D Gaussian splatting, excels in generating high-quality 3D artistic scenes from textual descriptions. Beyond quantitative metrics, ART3D significantly contributes to the intersection of AI and art by providing a novel solution for creating visually appealing 3D scenes.

{
    \small
    \bibliographystyle{ieeenat_fullname}
    \bibliography{main}
}

\end{document}